\pdfoutput=1

\documentclass[11pt]{article}

\usepackage{emnlp2021}

\usepackage{times}
\usepackage{latexsym}

\usepackage[T1]{fontenc}

\usepackage[utf8]{inputenc}

\usepackage{microtype}

\usepackage{xspace}
\usepackage{multirow}
\usepackage{xspace,mfirstuc,tabulary}
\usepackage{booktabs}
\usepackage{graphicx}
\usepackage{amsmath,amssymb}
\usepackage{header}
\usepackage[inline]{enumitem}

\title{Unlocking Compositional Generalization in Pre-trained Models\\ Using Intermediate Representations}

\author{Jonathan Herzig\thanks{\; Work completed while interning at Google.}$\mathbf{^{*,1}}$, Peter Shaw\textsuperscript{2}, Ming-Wei Chang\textsuperscript{2},\\ \textbf{Kelvin Guu\textsuperscript{2}, Panupong Pasupat\textsuperscript{2}, Yuan Zhang\textsuperscript{2}} \\
\\ $^1$Tel-Aviv University \quad $^2$Google Research\\
\texttt{\large jonathan.herzig@cs.tau.ac.il}\\ \texttt{\large\{petershaw,mingweichang,kguu,ppasupat,zhangyua\}@google.com}
\\
}

\begin{document}
\maketitle
\begin{abstract}
Sequence-to-sequence (seq2seq) models are prevalent in semantic parsing, but have been found to struggle at out-of-distribution compositional generalization. While specialized model architectures and pre-training of seq2seq models have been proposed to address this issue, the former often comes at the cost of generality and the latter only shows limited success.
In this paper, we study the impact of intermediate representations on compositional generalization in pre-trained seq2seq models, without changing the model architecture at all, and identify key aspects for designing effective representations. 
Instead of training to directly map natural language to an executable form, we map to a reversible or lossy {intermediate representation} that has stronger structural correspondence with natural language.
The combination of our proposed intermediate representations and pre-trained models is surprisingly effective, where the best combinations obtain a new state-of-the-art on CFQ (+14.8 accuracy points) and on the template-splits of three text-to-SQL datasets (+15.0 to +19.4 accuracy points).
This work highlights that intermediate representations provide an important and potentially overlooked degree of freedom for improving the compositional generalization abilities of pre-trained seq2seq models.

\end{abstract}

\section{Introduction}

\emph{Compositional generalization} is the desired ability for a semantic parser to generalize to new combinations of program components, where each component was seen at training time, but where the particular combination is out-of-distribution. For example, a parser trained on questions such as \utt{Rivers crossing New York} and \utt{What states border Texas?} should generalize to \utt{Rivers that cross states bordering Washington} at test time.

While sequence-to-sequence (seq2seq) models dominate semantic parsing \cite{jia-liang-2016-data,dong-lapata-2016-language,wang-etal-2020-rat}, previous work found that they perform poorly on evaluation that requires compositional generalization
~\cite{finegan-dollak-etal-2018-improving,lake2018generalization,keysers2019measuring}.
Both new architectures~\cite[\emph{inter alia}]{li2019compositional,lake2019compositional,nye2020learning,chen2020compositional} and general-purpose pre-trained seq2seq models such as T5~\cite{raffel2020exploring} have shown improvements on some evaluations of compositional generalization, but strong performance in general remains a significant challenge~\cite{shaw2020compositional,furrer2020compositional}.

\begin{figure}[t]
    \centering
    \includegraphics[width=\columnwidth]{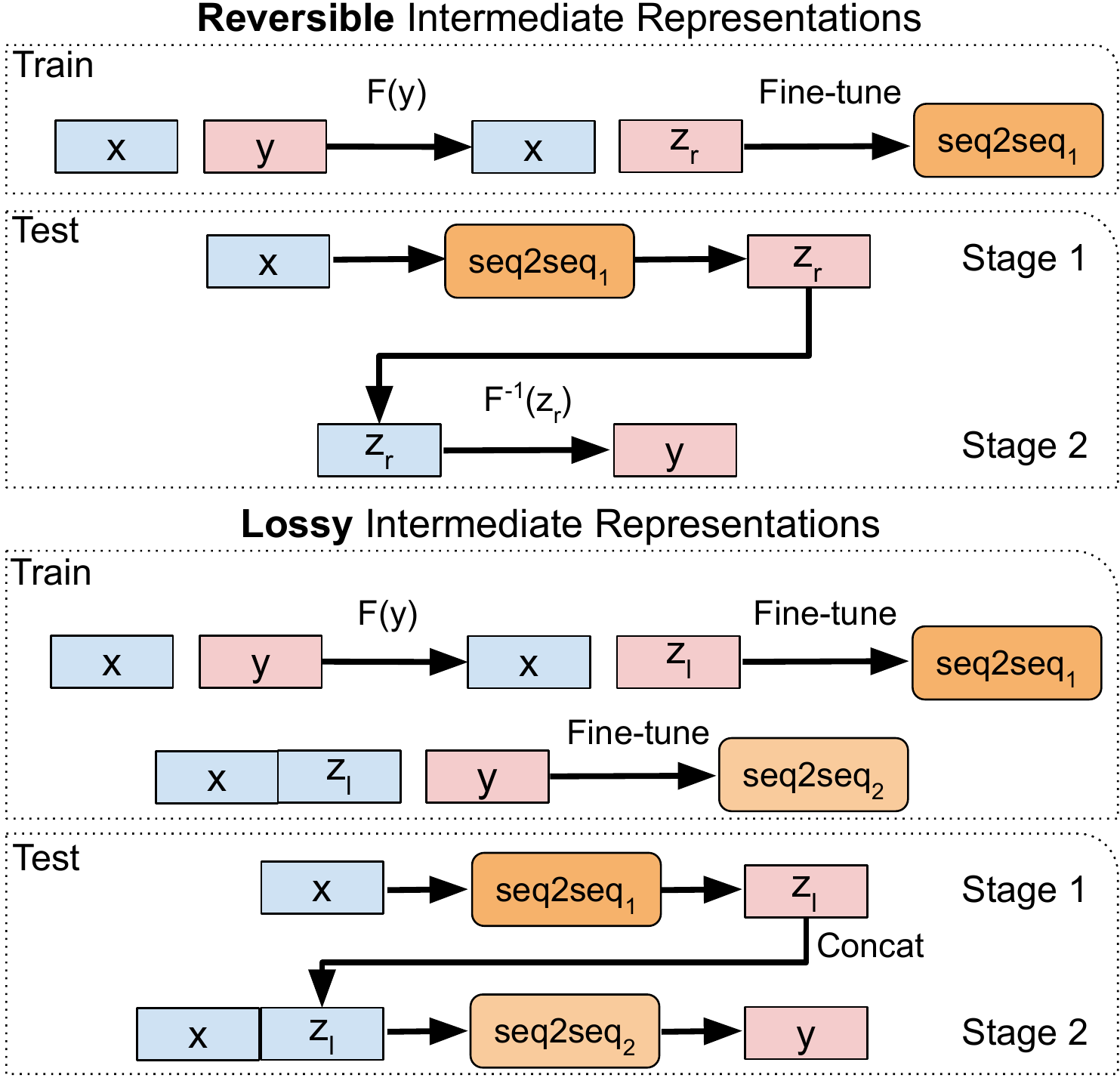}
    \caption{Our framework for parsing utterances ($x$) into programs ($y$) through reversible ($z_r$) and lossy ($z_l$) intermediate representations using seq2seq models.}
    \label{fig:transformation_intro_fig}
\end{figure}

In this paper we posit that pre-trained seq2seq models struggle with compositional generalization in part due to a low structural correspondence between the natural language and its meaning representation. Thus,
instead of training to directly map natural language to an executable form, we map to an \emph{intermediate representation} designed to increase the structural correspondence with natural language: for example, omitting elements of the parse that cannot be easily aligned to natural language, and adding structural cues such as brackets to indicate nested scopes. Since the intermediate form is no longer executable and may not even contain all details necessary for execution, we then apply a second stage to convert the intermediate representation into an executable parse. This is done using either deterministic transformations or a second seq2seq model that conditions on both the intermediate representation and the original natural language utterance, as illustrated in Figure~\ref{fig:transformation_intro_fig}.
Notably, we find that our intermediate representations are specifically effective when combined with \emph{pre-training}, suggesting that they help unlock information acquired during the pre-training process.

Our approach is closely related to previous work on coarse-to-fine decoding~\cite{dong-lapata-2018-coarse}, intermediate representations~\cite{guo-etal-2019-towards,suhr-etal-2020-exploring} and seq2seq2seq~\cite{baziotis2019seq}.
Our main novelty is the focus on the context of compositional generalization and our model agnostic design for two-stage decoding. Previous methods mostly targeted in-distribution generalization or did not find significant benefits in the compositional generalization setup~\cite{furrer2020compositional}.
Therefore, we believe our paper is a first thorough attempt to explore how intermediate representations can be combined with pre-training to improve compositional generalization.

The contributions of our paper are:\footnote{Our code is available at \url{https://github.com/google-research/language/tree/master/language/compir}.}

\begin{itemize}[leftmargin=*,itemsep=0pt]
    \item We study the effect of intermediate representations on the compositional generalization in pre-trained seq2seq models, and identify key aspects for designing effective representations.
    \item 
    We show that a well-designed intermediate representation is often synergistic with pre-training: when both are used together, the gains are bigger than the sum of each individually.
    \item We show that the best combinations of our intermediate representations with pre-trained seq2seq models obtain new state-of-the-art results on CFQ~\cite{keysers2019measuring} (+14.8 accuracy) and the template-splits of three text-to-SQL datasets~\cite{finegan-dollak-etal-2018-improving} (+15.0 to +19.4 accuracy), outperforming previous work by a large margin while maintaining competitive performance on the i.i.d. (random) splits.
\end{itemize}
\section{Review: Semantic Parsing Formalisms}
\label{sec:background}

\begin{figure*}[t]
    \centering
    \includegraphics[width=\textwidth]{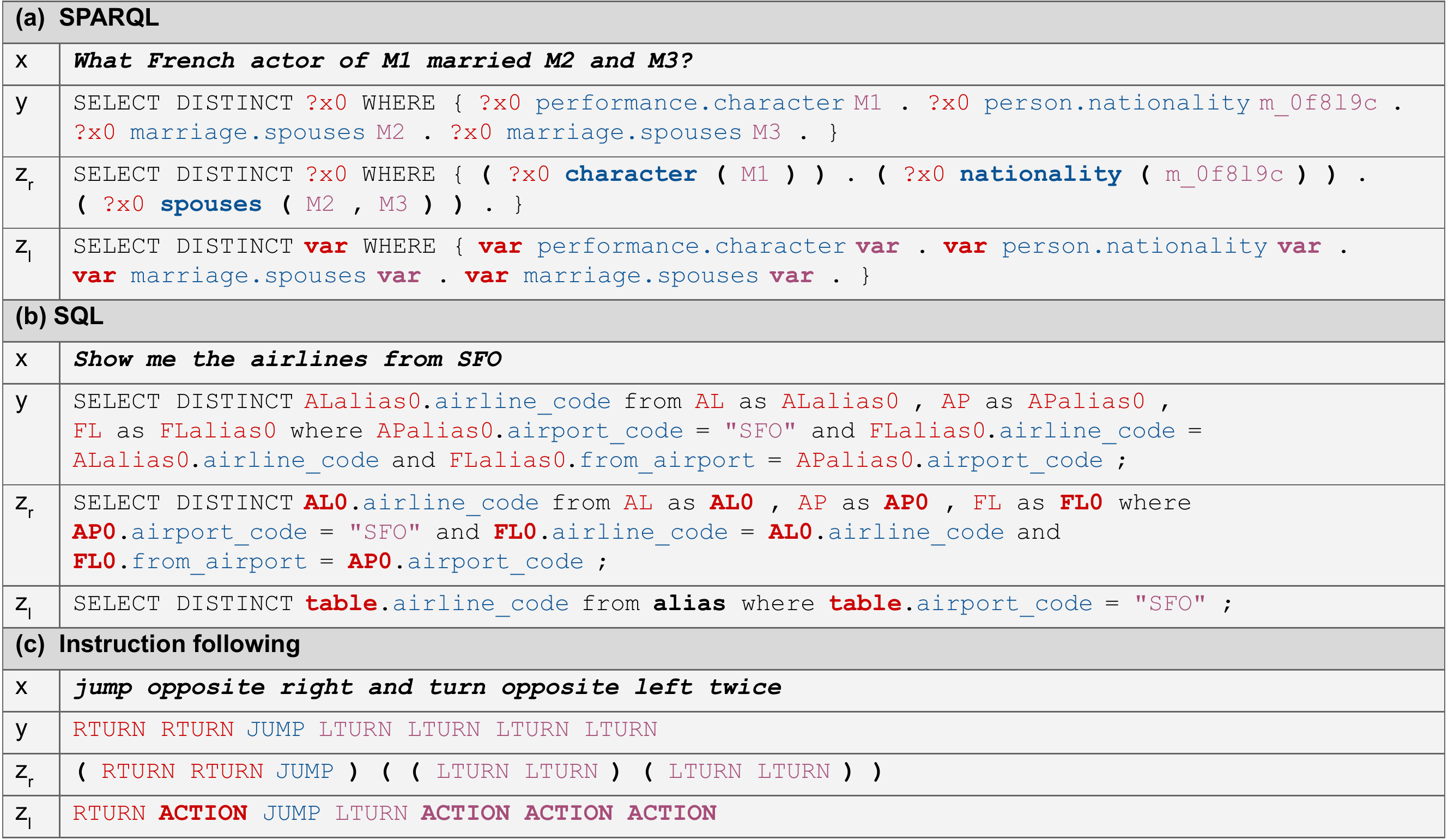}
    \caption{Examples for the different formalisms, of an utterance ($x$), program ($y$), reversible intermediate representation ($z_r$) and lossy intermediate representation ($z_l$). For each formalism, tokens with the same color share their semantic role. Tokens in $z_r$ and $z_l$ that are modified w.r.t. $y$ are in bold. We abbreviate original SPARQL relations, and also abbreviate the SQL table names \texttt{airline}, \texttt{airport} and \texttt{flight} to \texttt{AL}, \texttt{AP} and \texttt{FL}, respectively.}
\label{fig:dataset_examples}
\end{figure*}

We briefly describe the semantic parsing formalisms we explore in this work.
Figure~\ref{fig:dataset_examples} shows
an example utterance $x$ and program $y$ for each formalism.
We first experiment with \textbf{SPARQL} programs from \cfq{}~\cite{keysers2019measuring}.
Each SPARQL program contains a set of conjuncts, each of which consists of a subject, relation, and object.
For example, the conjunct \texttt{?x0 people.person.nationality m\_0f8l9c} limits the possible values for the variable \texttt{?x0} to people of French nationality.
Second, we experiment with \textbf{SQL} programs, canonicalized by \newcite{finegan-dollak-etal-2018-improving} to a consistent writing style where, e.g., table aliases are in the form \texttt{<TABLE NAME>alias<N>} for the \texttt{N}th use of the same table in one program. Finally, we consider the \textbf{instruction following} formalism in \scan{} \cite{lake2018generalization}, where natural language commands (e.g., \utt{jump twice}) are mapped to action sequences (e.g., \texttt{JUMP JUMP}).

For all of these formalisms, there can be a significant degree of structural mismatch between an utterance and its corresponding program. We hypothesize that this structural mismatch contributes to poor generalization on out-of-distribution compositional examples, even for pre-trained models.

\section{Intermediate Representations}
\label{sec:io_mod}

We study \emph{intermediate representations} (IRs) 
to improve compositional generalization for seq2seq models. For a program $y$, an IR can be defined as $z = \mathcal{F}(y)$ where $\mathcal{F}$ is a deterministic transformation function. Example IRs ($z$) are shown in Figure~\ref{fig:dataset_examples}. As we will explain in \S\ref{ssec:ir_design}, we design IRs to simplify programs and increase their structural correspondence with the input utterance.\footnote{Note that $z$ is not necessarily shorter than the original program $y$, as $z$ could carry additional structural cues.} Additionally, to make use of existing pre-trained models, our transformations $\mathcal{F}$ are  model-agnostic and do not require any architecture change to the model.

As shown in Figure~\ref{fig:transformation_intro_fig},
our framework for incorporating IRs when parsing utterances $x$ into programs $y$ consists of two stages.
First, instead of predicting $y$ directly from the utterance $x$, our seq2seq model $\texttt{Seq2Seq}_1: x\tox z$ is trained to predict the IR $z$.
Second, we map the predicted $z$ to a program $y$, using one of the two methods depending on the reversibility of the transformation $\mathcal{F}$. 
If $\mathcal{F}$ is reversible,
then the IR $z$ is a \emph{reversible IR} and contains all information to reconstruct the program $y$. In this case, we can simply call the inverse transformation $\mathcal{F}^{-1}: z \to y$ to produce the final prediction (Figure~\ref{fig:transformation_intro_fig}, top).
If $\mathcal{F}$ is irreversible, the IR is said to be \emph{lossy}. In this case we train an additional seq2seq model $\texttt{Seq2Seq}_2: (x\sep z)\tox y$ that predicts a program $y$ in the original formalism conditioned on both $x$ and $z$
(Figure~\ref{fig:transformation_intro_fig}, bottom).
This process is similar to coarse-to-fine decoding~\cite{dong-lapata-2018-coarse} if the lossy IR is designed to carry properties of a coarse sketch.

\subsection{Intermediate Representation Design}
\label{ssec:ir_design}

Our specific adaptation of each transformation $\mathcal{F}$ is guided by common errors we encountered in preliminary experiments with seq2seq on the original utterance-program pairs. Overall, based on our observations, we synthesize the following principles for constructing intermediate representations.

First, \emph{reduce the mismatch between the formal and natural language.} This can be implemented in two ways: omitting program elements that cannot be easily aligned to the utterance (e.g., the \texttt{FROM} clause in SQL), and rearranging the program structure (e.g., by grouping conjuncts in SPARQL).
Second, \emph{increase structural similarities among programs.} For example, anonymizing elements such as entities and variables in SPARQL can lead to different programs in the train and test sets having an identical IR, which improves generalization.
Finally, \emph{induce hierarchical structure into programs to indicate scoping.} For instance, we add brackets for different program components in \scan{}.

\subsection{Reversible Intermediate Representation}
\label{ssec:fm}

We adapt the following reversible IRs for each formalism ($z_r$ in Figure \ref{fig:dataset_examples}).

\paragraph{SPARQL}
A mismatch between SPARQL programs and utterances often occurs when relations are expressed in a distributive manner in language. For example, in Figure \ref{fig:dataset_examples} (a), while \texttt{marriage.spouses} appears twice in the program $y$, it is manifested once in the utterance $x$ (\utt{\textbf{married} M2 and M3}). To alleviate this mismatch, we group conjuncts that share the same subject and relation; for example, given a SPARQL query with triples
$\{(x_0,r_1,e_1)$, $(x_0,r_1,e_2)$, $(x_0,r_2,e_1)$, $(x_0,r_2,e_2)\}$, we modify it to $\{(x_0,r_1,(e_1, e_2))$, $(x_0,r_2,(e_1,e_2))\}$.
We further shorten the IR by truncating relation names of the form \texttt{<$r_{\text{prefix}}$><ns:><$r_{\text{suffix}}$>} to \texttt{<$r_{\text{suffix}}$>} while ensuring the final relation name is still unique. Finally, to induce additional hierarchical structure, we add brackets around each conjunct.
We find this IR to be more effective than those proposed by previous work \cite{furrer2020compositional,guo2020hierarchical}. Later, in \S\ref{ssec:ir_ablations}, we find each of our design decisions to individually assist performance.

\paragraph{SQL}

Previous work has proposed various IRs for SQL, such as SemQL~\cite{guo-etal-2019-towards}, SQL\textsuperscript{\texttt{UF}}\xspace~\cite{suhr-etal-2020-exploring}, and an extension of relational algebra~\cite{rubin2020smbop}. However, these
IRs are primarily developed for the Spider dataset~\cite{yu2018spider}, and their conversion procedures make various assumptions that can limit their applicability to other datasets. For example, unlike Spider, datasets such as ATIS contain SQL queries with self joins and multiple foreign key relations between a given pair of tables. Consequently, \citet{suhr-etal-2020-exploring} note that less than $20\%$ of the queries in ATIS can be successfully converted to their intermediate representations, SQL\textsuperscript{\texttt{UF}}\xspace.

Therefore, we propose a simple IR aimed at omitting tokens in the program that do not align to any phrase in the utterance. By convention, the datasets we study use table aliases of the form \texttt{<TABLE NAME>alias<N>} for the \texttt{N}th use of the same table in a query. For the reversible IR, we simply remove the \texttt{alias} token, such that table alias names take the form of \texttt{<TABLE NAME><N>}. We will further simplify the program using lossy IRs (\S\ref{ssec:c2f}).

\paragraph{Instruction Following} 
Programs in \scan{} consist of a sequence of actions with no explicit hierarchical (e.g., tree) structure that could indicate the scope of actions in the program.
This lack of structure is conveyed by common errors we encounter when decoding reoccurring actions. For example, \utt{turn opposite left twice} in Figure \ref{fig:dataset_examples} (c) should be mapped to four consecutive \texttt{LTURN} actions. In this case the model often wrongly decodes \texttt{LTURN} more or less times than required. 
We induce a hierarchical structure into the reversible IR by adding brackets around repeated program components (e.g., each of the two occurrences that \utt{turn opposite left twice} maps to) and around complex actions (e.g., the program component that \utt{jump opposite right} maps to), which could assist the model in mapping language phrases to actions.\footnote{We use a synchronous CFG to add this bracketing, so for SCAN the transformation $\mathcal{F}$ is also a function of $x$.}

\subsection{Lossy Intermediate Representations}
\label{ssec:c2f}
The IRs suggested in \S\ref{ssec:fm} are required to be fully reversible such that we can recover the executable program in the original formalism without information loss.
However, we find lossy IRs that omit or anonymize program components to be desired. For example, table names in SQL are frequently absent from utterances, such as \texttt{airport} and \texttt{flight} in Figure~\ref{fig:dataset_examples}(b), and thus predicting them correctly can be difficult.

Figure \ref{fig:dataset_examples} illustrates examples of the lossy IRs ($z_l$) we propose.
For SPARQL, we anonymize entities and variables, replacing them with the placeholder \texttt{var}, which increases the similarity between different $z_l$ instances. For SQL, we adapt $z_l$ to only contain components that are frequently manifested in the utterance (e.g., column names and values), which reduces the mismatch between utterances and their IR $z_l$. Particularly, we omit the \texttt{FROM} clause, mask table names, and omit conditions that are only relevant for joining tables. For instruction following, we anonymize repeated single actions, except the first one. Formally, we replace strings $\texttt{C}^n$ (action \texttt{C} repeating $n>1$ times) with $\texttt{C}\;\texttt{A}^{n-1}$, where \texttt{A} represents an anonymized action. This IR increases the similarity of different $z_l$ instances and targets common errors for \scan{} (\S\ref{ssec:fm}). 

\paragraph{Using lossy IRs}
We experiment with two methods for predicting and utilizing lossy IRs.
The first method is {\em direct prediction},
where we use $\texttt{Seq2Seq}_1: x \tox z_l$ to directly produce the IR $z_l$, and then apply $\texttt{Seq2Seq}_2: (x\sep z_l) \tox y$ to predict the final program.
Here, we hypothesize that mapping to IRs is an easier learning problem for  $\texttt{Seq2Seq}_1$, and in addition, that the joint encoding of the utterance $x$ and the IR $z_l$ provides a rich context for  $\texttt{Seq2Seq}_2$.

To isolate the two hypotheses, we experiment with a second method, {\em indirect prediction}, where $\texttt{Seq2Seq}_1$ is trained to predict the original program instead of an IR. From the prediction $y^* = \texttt{Seq2Seq}_1(x)$, we apply the lossy transformation to create the IR $z_l = \mathcal{F}(y^*)$ before applying $\texttt{Seq2Seq}_2: (x\sep z_l) \tox y$.

\section{Experiments}
\label{sec:experiments}

\subsection{Datasets}
\label{ssec:datasets}

We evaluate performance on (1) \cfq{}~\cite{keysers2019measuring}; (2) three text-to-SQL datasets curated by \citet{finegan-dollak-etal-2018-improving}, including \geo{}~\cite{zelle-mooney-1996-learning}, \atis{}~\cite{dahl-etal-1994-expanding}, and \scholar{}~\cite{iyer-etal-2017-learning};
and (3) \scan{}~\cite{lake2018generalization}.
The formalism for each dataset is described in \S\ref{sec:background}, and the dataset sizes are given in Table \ref{tab:dataset_size} in the Appendix.

We consider multiple dataset splits that aim to assess compositional generalization.
For \cfq{} and \scan{}, we use the \emph{Maximum Compound Divergence} (MCD) splits~\cite{keysers2019measuring}, which are generated by making the distributions of compositional structures in the train and test sets as divergent as possible.
For the text-to-SQL datasets, we use \emph{template} splits~\cite{finegan-dollak-etal-2018-improving}, which ensure that the train and test set contain distinct SQL query ``templates'' (constructed by replacing values in the SQL queries with anonymized placeholders).
Finally, for SCAN, we use two additional splits from \citet{lake2018generalization}: the \emph{length} split, which requires generalization to longer sequences, and the \emph{turn left} split, where the ``turn left'' command is recombined with other elements of the training set in novel ways at test time.

\subsection{Experimental Setup}
Unless stated otherwise, each seq2seq in our model is a pre-trained T5 model~\cite{raffel2020exploring} fine-tuned on appropriate input-output pairs (e.g., $(x, z_r = \mathcal{F}(y))$ pairs for $\texttt{Seq2Seq}_1: x\tox z_r$).
We only tune the leaning rate for each dataset on the dev set, considering the values $[1e^{-3}, 5e^{-4}, 1e^{-4}]$. For \scan{}, as no dev set is available, we tune the learning rate on the \textit{around right} split~\cite{loula-etal-2018-rearranging}, which we use as a held-out set. We use batch size of 128 and fine-tune all models for $30K$ steps. We evaluate our models on exact-match accuracy. 
For \cfq{}, we post-process predicted programs by sorting conjuncts alphabetically and removing duplicate conjuncts, similar to \citet{guo2020hierarchical}.

\subsection{Pilot Study}

We first conduct a pilot study where we experiment with the different IRs in \S\ref{sec:io_mod} on all compositional splits to evaluate their potentials.
Against the no-transformation baseline ($x \tox y$),
we consider using the
reversible IR (RIR: $x \tox z_r \to y$) and the
lossy IR with either direct prediction (\LIRdir: $x \tox z_l \tox y$) or indirect prediction (\LIRind: $x \tox y^* \to z_l \tox y$)\footnote{Note that the predicted $y^*$ from the first seq2seq can be different from the final prediction $y$, as described in \S\ref{ssec:c2f}.}.
As LIR and RIR are independent, we also experiment with pipelining them together (\LIRdir+RIR: $x \tox z_{l,r} \tox z_r \to y$ and \LIRind+RIR: $x \tox z^*_r \to z_{l,r} \tox z_r \to y$, where $z_{l,r}$ is the result of applying both the reversible and lossy transformations).
We use T5-base as the seq2seq model.

\begin{table*}[t]
\centering
\resizebox{0.92\textwidth}{!}{
\begin{tabular}{@{}lccccccccccc@{}}
\toprule
\multirow{2}{*}{Model} & \multicolumn{3}{c}{CFQ} & \multicolumn{3}{c}{Text-to-SQL}                & \multicolumn{5}{c}{SCAN} \\
\cmidrule(lr){2-4}
\cmidrule(lr){5-7}
\cmidrule(lr){8-12}
                          & MCD1   & MCD2   & MCD3  & \atis{}   & \geo{}   & \scholar{} & Length & Turn Left & MCD1 & MCD2 & MCD3   \\
\midrule
Baseline                  & 58.5   & 27.0   & 18.4  & 32.9   & 79.7       & 18.1      & 14.5   & 66.1      & 15.2  & 14.3  & 10.6 \\
RIR                        & 86.3  & 49.1   & 46.8  & 36.3   & 81.3       & 19.4      & 54.7   & 100.0       & 100.0   & 100.0   & 75.3 \\
\LIRdir                      & 48.1   & 40.3   & 35.3  & 44.4   & 83.5       & 20.6      & 14.2 &	83.5 &	15.7 &	13.2 &	17.5   \\
\LIRdir+RIR                        & 72.5   & 61.1   & 51.2  & 47.8   & 83.0       & 20.0      & 56.4   & 100.0     & 100.0 & 100.0 & 75.1 \\
\LIRind                     & 57.6   & 41.4   & 34.7  & 38.3   & 80.8       & 16.5      & 13.5   & 65.7      & 15.0  & 13.8  & 10.6 \\
\LIRind+RIR                       & 85.8   & 64.0   & 53.6  & 41.5   & 81.9       & 16.5      & 54.4   & 100.0     & 100.0 & 100.0 & 75.0 \\
\bottomrule
\end{tabular}}
\caption{Results on the test set for all approaches and all compositional splits with T5-base.}
\label{tab:results_explore}
\end{table*}

The results in Table~\ref{tab:results_explore} show that for \cfq{}, RIR improves baseline performance significantly, from an average of 34.6 to 60.8.
Combining RIR with \LIRind{} further boosts average performance on the MCD splits to 67.8. While \LIRdir{} performs much better than the baseline on average, it lags behind other transformations on MCD1, e.g., 9.5 point worse than \LIRind{} (48.1 vs 57.6). A closer look shows that exact-match accuracy of $z_l$ predicted by $\texttt{seq2seq}_1$ on MCD1 is only 47.2, suggesting that anonymizing variables and entities might hide relevant information that could assist $\texttt{seq2seq}_1$ to predict the correct lossy IR.

For text-to-SQL datasets, even our simple RIR, where some tokens are omitted from the program, yields improvements across all datasets. Combining RIR with \LIRdir{} further achieves significant improvements over the baseline, especially for ATIS (from 32.9 to 47.8).

On \scan{}, RIR significantly improves baseline accuracy, achieving perfect accuracy for the \emph{turn left}, MCD1 and MCD2 splits. On the \emph{length} split, RIR yields a boost of 40 accuracy points even though generalizing to longer programs is a known challenge for seq2seq models~\cite{newman-etal-2020-eos}. 
This shows that by injecting a small amount of additional information about the hierarchical structure of the output programs, we can outperform previous results for seq2seq models, and match the results of specialized architectures such as LANE~\cite{liu2020compositional} across most splits.
As for LIRs, except for \LIRdir{} we do not observe major improvements over the baseline and RIR. This is reasonable, as program elements in \scan{} have overall close alignment to phrases in the utterance.

\begin{table}[t]
\centering
\resizebox{0.87\columnwidth}{!}{
\begin{tabular}{@{}lcccc@{}}
\toprule
Model                  & MCD1 & MCD2 & MCD3 & Ave. \\
\midrule
LSTM+A~ $\clubsuit$                 & 28.9 & 5.0  & 10.8 & 14.9 \\
Transformer~ $\clubsuit$              & 34.9 & 8.2  & 10.6 & 17.9 \\
Univ. Trans.~ $\clubsuit$             & 37.4 & 8.1  & 11.3 & 18.9 \\
Evol. Trans.~ $\diamondsuit$           & 42.4 & 9.3  & 10.8 & 20.8 \\
IBT~ $\spadesuit$                    & 64.8 & 57.8 & 64.6 & 62.4 \\
HPD~ $\heartsuit$                    & 79.6 & 59.6 & 67.8 & 69.0 \\
\midrule
\midrule
Baseline (T5-base)     & 58.5 & 27.0 & 18.4 & 34.6 \\
Baseline (T5-large)    & 65.1 & 32.3 & 25.4 & 40.9 \\
Baseline (T5-3B)       & 65.0 & 41.0 & 42.6 & 49.5 \\
RIR (T5-base)           & 86.3 & 49.1 & 46.8 & 60.8 \\
RIR (T5-large)          & \textbf{88.7} & 62.2 & 57.1 & 69.3 \\
RIR (T5-3B)             & \textbf{88.7} & 72.6 & 63.5 & 75.0 \\
\LIRind+RIR (T5-base)     & 85.8 & 64.0 & 53.6 & 67.8 \\
\LIRind+RIR (T5-large)    & 88.6 & 79.2 & 72.7 & 80.2 \\
\LIRind+RIR (T5-3B)       & 88.4 & \textbf{85.3} & \textbf{77.9} & \textbf{83.8} \\
\bottomrule
\end{tabular}}
\caption{CFQ test set results for different model sizes and in comparison with previous work: $\clubsuit$~\cite{keysers2019measuring}, $\diamondsuit$~\cite{furrer2020compositional}, $\spadesuit$~\cite{guo2021revisiting} and $\heartsuit$~\cite{guo2020hierarchical}.
}
\label{tab:cfq_extended}
\end{table}

\subsection{Main Results}

Following our pilot study, we further experiment with the most promising IRs on \cfq{} and the text-to-SQL datasets, and compare performance across different model capacities (base, large and 3B).

\begin{table}[t]
\centering
\resizebox{0.86\columnwidth}{!}{
\begin{tabular}{@{}lccc@{}}
\toprule
Model                  & \atis{} & \geo{} & \scholar{} \\
\midrule
Seq2Seq~ $\clubsuit$                & 32.0 & 20.0     & 5.0     \\
GECA~ $\diamondsuit$                   & 24.0 & 52.1     & -       \\
Seq2Seq~ $\spadesuit$                & 28.0 & 48.5     & -       \\
Transformer~ $\spadesuit$            & 23.0 & 53.9     & -       \\
Seq2Seq+ST~ $\spadesuit$             & 29.1 & 63.6     & -       \\
Transformer+ST~ $\spadesuit$         & 28.6 & 61.9     & -       \\
\midrule
\midrule
Baseline (T5-base)     & 32.9 & 79.7     & 18.1    \\
Baseline (T5-large)    & 31.4 & 81.9     & 17.5    \\
Baseline (T5-3B)       & 29.7 & 79.7     & 16.2    \\
\LIRdir+RIR (T5-base)       & \textbf{47.8} & \textbf{83.0}     & 20.0    \\
\LIRdir+RIR (T5-large)      & 43.2 & 79.7     & \textbf{22.0}    \\
\LIRdir+RIR (T5-3B)         & 28.5 &	75.8	&  12.4 \\ \bottomrule
\end{tabular}}
\caption{Text-to-SQL test set results on the template splits, for different model sizes and in comparison with previous work: $\clubsuit$~\cite{finegan-dollak-etal-2018-improving}, $\diamondsuit$~\cite{andreas-2020-good}, and $\spadesuit$~\cite{zheng2020compositional}.}
\label{tab:sql_extended}
\end{table}

The CFQ results are in Table~\ref{tab:cfq_extended}.
In line with~\citet{furrer2020compositional}, we find that our T5 baseline already performs better than general seq2seq architectures with no pre-training, including LSTM with attention~\cite{bahdanau2015neural} and different Transformer variants~\cite{NIPS2017_3f5ee243,dehghani2018universal,so2019evolved}.
Our IRs then significantly improve upon the baseline performance, and this improvement compounds with model capacity. With T5-large, simply using RIR already yields 0.3 accuracy points over HPD~\cite{guo2020hierarchical}, the current state-of-the-art that utilizes a specialized architecture tailored for CFQ. We further note that an IR proposed by ~\citet{furrer2020compositional} for CFQ, that differently than ours, groups by subjects and objects, was only found to improve a T5 baseline by 1.2 points.
Jointly applying RIR and \LIRind{} gives further improvements that also compound with model capacity. With T5-large and T5-3B, the models surpass state-of-the art, yielding accuracy of 80.2 and 83.8, respectively.

Text-to-SQL results are in Table~\ref{tab:sql_extended}. We compare with general seq2seq architectures (Seq2Seq and Transformer), GECA~\cite{andreas-2020-good} — a data augmentation method evaluated by \citet{zheng2020compositional}, and with ST (semantic tagging) which targets compositional generalization. We further find that for T5-base, the no-transformation baseline is already on-par with (\atis{}) or surpasses (\geo{} and \scholar{}) the state-of-the-art. \LIRdir+RIR yields additional gains and achieves new state-of-the-art on all three datasets. For both our T5 baseline and \LIRdir+RIR, further increasing model capacity beyond T5-base does not give further improvements, which is consistent with previous work on similar tasks with small train set sizes~\cite{shaw2020compositional,furrer2020compositional}.

\begin{figure}[t]
    \centering
    \definecolor{myblue}{rgb}{0.2588,0.5215,0.9568}
\definecolor{myred}{rgb}{0.9176,0.2627,0.2078}
\begin{tikzpicture}
\begin{axis}[
    ybar=1pt,       
    height=10em,
    width=3.4in,
    bar width=5pt,
    enlarge x limits=0.05,  
    clip=false,     
    ymin=0, ymax=100,
    axis lines*=left,
    y axis line style={
        draw=none,
    },
    ymajorgrids=true,
    xtick={
        0,1,2,3,4,
        5.5,6.5,7.5,8.5,9.5,10.5,11.5
    },
    xticklabels={
        CFQ,
        \atis{},
        \geo{},
        \scholar{},
        SCAN,
        CFQ MCD avg,
        \atis{},
        \geo{},
        \scholar{},
        SCAN Length,
        SCAN Turn Left,
        SCAN MCD avg,
    },
    xticklabel style={
        rotate=45,
        anchor=east,
        font=\tiny,
        xshift=4pt,
    },
    yticklabel style={
        font=\tiny,
    },
    tick style={draw=none},
    legend style={
        legend columns=-1,
        at={(0.5,1.25)},
        anchor=south,
        draw=none,  
        font=\footnotesize,
    },
]
\draw[line width=1pt,gray]
    (axis cs: 4.75, 0) -- (axis cs: 4.75, 125);
\node[anchor=south]
    at (axis cs: 2.2, 100) {\textbf{i.i.d.\ splits}};
\node[anchor=south]
    at (axis cs: 8.5, 100) {\textbf{compositional splits}};
\addplot[draw=myblue,fill=myblue]
coordinates {
    (0, 99.5)
    (1, 58.6)
    (2, 80.6)
    (3, 72.9)
    (4, 100.0)
    (5.5, 34.6)
    (6.5, 32.9)
    (7.5, 79.7)
    (8.5, 18.1)
    (9.5, 14.5)
    (10.5, 66.1)
    (11.5, 13.4)
    };
\addplot[draw=myred,fill=myred]
coordinates {
    (0, 99.4)
    (1, 59.5)
    (2, 81.0)
    (3, 75.2)
    (4, 100.0)
    (5.5, 67.8)
    (6.5, 47.8)
    (7.5, 83.5)
    (8.5, 20.6)
    (9.5, 56.4)
    (10.5, 100.0)
    (11.5, 91.8)
    };
\legend{Baseline, Best IR}
\end{axis}
\end{tikzpicture}
    \caption{Compared to Baseline (T5-base), the best IR of each split maintains the baseline accuracy for i.i.d.\ splits while giving large gains for compositional splits.}
    \label{fig:baseline_vs_ir_bar_chart}
\end{figure}
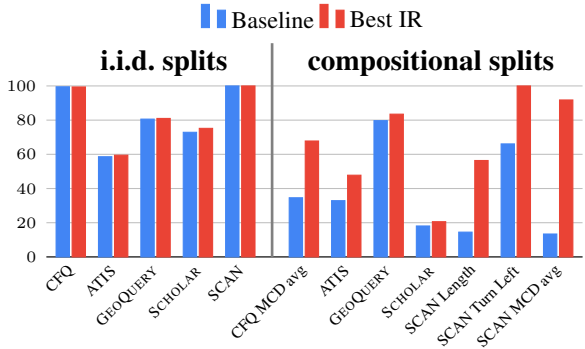

\subsection{Performance on i.i.d.\ Splits}

While our proposed IRs substantially improve the performance of T5 on compositional splits, we wish to verify they do not hurt performance on i.i.d.\ splits. To this end, we test our approaches with T5-base on the \emph{random} splits of \scan{} and \cfq{}, and on the standard i.i.d.\ splits of the text-to-SQL datasets.
As shown in Figure~\ref{fig:baseline_vs_ir_bar_chart} (see Table~\ref{tab:iid_results} for full results), we find that IRs indeed maintain the baseline accuracy on these i.i.d.\ splits.
\section{Analysis}
\label{sec:analysis}

\subsection{Interaction with Pre-Training}
\label{ssec:pretraining}

\begin{table}[t]
\centering
\resizebox{1.0\columnwidth}{!}{
\begin{tabular}{@{}lcccc@{}}
\toprule
Model                  & \multicolumn{2}{c}{CFQ} & \multicolumn{2}{c}{Text-to-SQL} \\
\midrule
T5-small w/o pre-training & 20.8 & & 33.7 & \\
\;\;+Pre-training    & 28.0 & \textcolor{forestgreen}{+7.2} & 41.8 & \textcolor{forestgreen}{+8.1} \\
\;\;+IRs & 22.6 & \textcolor{forestgreen}{+1.8} & 21.9 & \textcolor{red}{-11.8} \\
\;\;+Pre-training+IRs & 47.9 & \textcolor{forestgreen}{+27.1} & 46.5 & \textcolor{forestgreen}{+12.8}\\
\bottomrule
\end{tabular}}
\caption{Impact of adding pre-training and intermediate representations (\LIRind+RIR for CFQ; \LIRdir+RIR for text-to-SQL) over a baseline model. The results are averaged over different test sets.}
\label{tab:pt_and_ir}
\end{table}

To further inspect whether improvements from IRs occur due to T5 pre-training, we fine-tune a T5-small model without loading the pre-trained weights. Table~\ref{tab:pt_and_ir} shows that for a model with no pre-training, IRs only give modest improvements or even hurt the accuracy. This suggests that our proposed IRs specifically assist T5 to unlock information it has acquired during pre-training.

\subsection{Ablation on Reversible IR}
\label{ssec:ir_ablations}

Results in \S\ref{sec:experiments} show that RIR has a large impact on compositional generalization, particularly for \cfq{}. To understand the impact of each design decision, we ablate aspects of our RIR for SPARQL on \cfq{}.
Table~\ref{tab:ir_ablate} shows that all ablations hurt performance.
The largest drop in performance (60.8 to 38.7) comes from removing the merging of conjuncts with shared relations and objects.

To see if RIR increases structural similarity between programs, we calculate the percentage of new structures (defined as the result of anonymizing entities and variables in $z_r$) that appear in the dev set with respect to the train set. 
We also calculate the average length (number of word-pieces) of programs in the dev set to see if RIR helps reduce program complexity.
Table~\ref{tab:ir_ablate} shows that performance correlates with having fewer novel structures and shorter programs.
This suggests that our design choices for RIR, discussed in \S\ref{ssec:ir_design}, contribute to compositional generalization.

\subsection{Ablation on Lossy IR}
\label{ssec:c2f_analysis}

We analyze two variations to our proposed usage of \LIRdir{}.
(1) \LIRcat, where instead of using two separate models for predicting the IR $z_l$ and the program $y$, we only use one model that predicts $z_l$ concatenated with $y$. This differs from \LIRdir{} and \LIRind{} in terms of model capacity and in how the model attends to the context when generating program tokens.
(2) VARified Baseline, where we use a single model to predict the program like in the baseline, but the generated program should have an additional \texttt{var} token before each variable and entity to indicate the similar role they share (e.g., \texttt{?x0 marriage.spouses M2} becomes \texttt{var ?x0 marriage.spouses var M2}).
This is to see if the usage of \texttt{var} in our LIR can be effective without explicitly predicting an IR.

\begin{table}
\centering
\resizebox{0.98\columnwidth}{!}{
\begin{tabular}{@{}lcccc@{}}
\toprule
Model           & \multicolumn{2}{c}{\begin{tabular}[c]{@{}c@{}}MCD \\ average\end{tabular}} & \begin{tabular}[c]{@{}c@{}}\% New \\ structures\end{tabular} & \begin{tabular}[c]{@{}c@{}}Ave. length \\ (word pieces)\end{tabular} \\
\midrule
Baseline           & 34.2                               &                                    & 91.7                                                        & 161      \\
\midrule
\midrule
RIR                 & 60.8                               &                                    & 80.9                                                        & 104      \\
\;\;-shorter relations   & 58.2                               & \textcolor{red}{-2.6}                               & 80.9                                                        & 128      \\
\;\;-brackets          & 57.2                               & \textcolor{red}{-3.6}                               & 80.9                                                        & 96      \\
\;\;-merge conjuncts & 38.7                               & \textcolor{red}{-22.1}                              & 91.7                                                        & 147     \\
\bottomrule
\end{tabular}}
\caption{CFQ dev accuracy for RIR ablations.}
\label{tab:ir_ablate}
\end{table}

\begin{table}[t]
\centering\small
\begin{tabular}{@{}lc@{}}
\toprule
Model choice & MCD average \\
\midrule
Baseline       & 34.2       \\
\LIRdir{}            & 41.9       \\
\LIRcat        & 30.8       \\
VARified Baseline        & 38.8       \\
LIR-ORACLE     & 80.7    \\
\bottomrule
\end{tabular}
\caption{CFQ dev accuracy for \LIRdir{} alternatives.}
\label{tab:c2f_analysis}
\end{table}

In addition, we run an oracle experiment LIR-ORACLE where we use the gold LIR as input to $\texttt{Seq2Seq}_2$ during inference, instead of using the prediction from $\texttt{Seq2Seq}_1$.

Table~\ref{tab:c2f_analysis} indicates that both \LIRcat{} and VARified Baseline achieve lower performance than LIR. However, while VARified Baseline still improves upon the baseline performance from 34.2 to 38.8, \LIRcat{} performs worse than the baseline. This could be partially explained by the fact that 7.4\% of the targets for \LIRcat{} exceed the maximal 512 tokens length after concatenation. Results for our oracle experiment, LIR-ORACLE, show that having the gold LIR during inference boosts performance from $41.9 \rightarrow{} 80.7$. This shows that the model can effectively utilize the LIR as context when generating the actual program.

\subsection{Example predictions}
\label{ssec:error_analysis}

\begin{figure}[t]
    \centering
    \includegraphics[width=\linewidth]{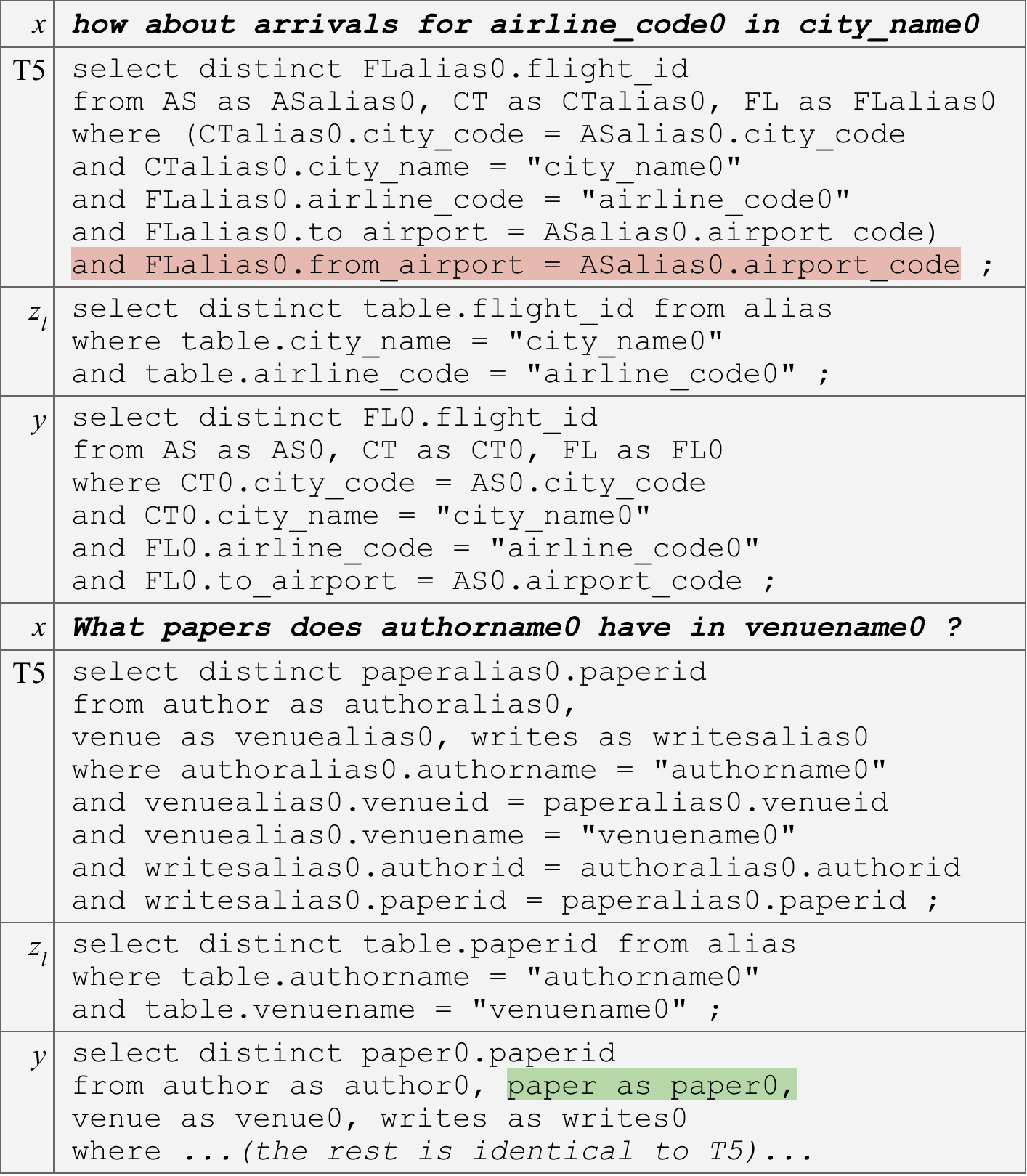}
    \caption{Example cases where \LIRdir+RIR produces correct programs whereas the baseline T5 does not. SQL table names were shortened here for brevity.}
    \label{fig:error_analysis_examples}
\end{figure}

We compare the predictions from the T5 baseline and \LIRdir+RIR (with T5-base) on the text-to-SQL development sets.
Figure~\ref{fig:error_analysis_examples} shows several examples where \LIRdir+RIR helps produce correct programs.

We can think of an SQL query as composed of two parts. The \emph{semantic} part are clauses that express information from the utterance, such as the \texttt{SELECT} clause and the \texttt{WHERE} filters (e.g., \texttt{airport = "SFO"}). This part tends to follow the compositional structure in the query. In contrast, the \emph{structural} part are additional clauses that make the query valid, such as the \texttt{FROM} clauses and the join clauses (e.g., \texttt{writes.paperid = paper.paperid}). The structural part usually depends on the semantic part, and generating them correctly requires schema reasoning.

The first example from Figure~\ref{fig:error_analysis_examples} shows how \LIRdir+RIR helps with compositional reasoning in the semantic part.
T5 generates the common pattern of having two cities and forces in the argument values.
By contrast, \LIRdir+RIR first predicts a shorter coarse program that only focuses on how the specified values are used as SQL filters. This makes the model less susceptible to blindly following common patterns found during training.

The second example shows how \LIRdir+RIR helps with schema reasoning in the structural part.
The baseline T5 generates the program from left to right, so it is more susceptible to creating structural inconsistencies. In contrast, \LIRdir+RIR first generates the coarse structure, which outlines the semantic part, then conditions on them to generate the final program. As such, the structural part is more likely to agree with the semantic part.

\subsection{Limitations}
\label{ssec:limitations}

Our approach suggests designing IRs for improving the compositional generalization abilities of the semantic parser. While we observe substantial gains in doing so, our approach requires customizing new IRs for each new formalism. While this entails manual work, it should be done only once per formalism (for example we are able to apply the same IRs for three different text-to-SQL datasets). Furthermore, the design choices needed for constructing IRs are similar to those needed when designing a coarse structure in coarse-to-fine decoding~\cite{dong-lapata-2018-coarse}, prompts when describing unsupervised tasks~\cite{NEURIPS2020_1457c0d6}, or task specific cloze-style patterns to help language
models understand a given task~\cite{schick2020exploiting}. Finally, our principles in~\S\ref{ssec:ir_design} may assist in reducing the time needed for designing IRs to a minimum.   
\section{Related Work}

\paragraph{Compositional Generalization}
In contrast to our work combining intermediate representation with pre-trained models, many other approaches have been pursued to improve compositional generalization in semantic parsing. These include new or modified model architectures~\cite{li2019compositional,russin2019compositional,gordon2020permutation,liu2020compositional,nye2020learning,chen2020compositional, zheng2020compositional, oren-etal-2020-improving,herzig2020span}, hybrid models~\cite{shaw2020compositional}, meta-learning~\cite{lake2019compositional}, and compositional data augmentation~\cite{andreas-2020-good}. Also, \citet{furrer2020compositional} compare pre-trained models vs specialized architectures for compositional generalization.

\paragraph{Intermediate Representations}
Unlike the formalisms we focus on in this work such as SQL, previous approaches to semantic parsing have often leveraged formal representations that were explicitly designed with correspondence to natural language in mind, such as FunQL~\cite{kate2005learning}, DCS~\cite{liang-etal-2011-learning}, and variants of typed lambda calculus~\cite{carpenter1997type,zettlemoyer2005learning}. Unlike our IRs, these formalisms typically require manual annotation.
~\citet{guo-etal-2020-benchmarking} compares performance of semantic parsers across several such formalisms as well as SQL.
Other work has focused on developing IRs that are domain independent~\cite{kwiatkowski-etal-2013-scaling,herzig-berant-2018-decoupling}.
We also discuss prior work~\cite{guo-etal-2019-towards,suhr-etal-2020-exploring,furrer2020compositional,guo2020hierarchical} developing reversible IRs for the formalisms we study in \S~\ref{ssec:fm}. 

The use of lossy IRs proposed in this work is closely related to the coarse-to-fine method of ~\citet{dong-lapata-2018-coarse}. However, their work did not consider pre-trained models, and they propose a specialized architecture. Such approaches commonly refer to the lossy IR as a \emph{sketch}. The use of sketches as an IR has also been explored for program synthesis~\cite{solar2008program,zhang2013automatically,nye2019learning}.
\section{Conclusion}

In this paper, we study simple yet effective strategies for constructing intermediate representations to improve compositional generalization abilities of pre-trained seq2seq models. We conduct extensive experiments on varied datasets and formalisms and our approaches consistently outperform state-of-the-art models by a large margin. We also demonstrate that our intermediate representations synergize well with pre-training, showing bigger gains than the sum of either alone when used together.
\section*{Acknowledgments}

We would like to thank Luheng He and Ohad Rubin for their useful comments. This work was completed in partial fulfillment for the PhD degree of the first author, which was also supported by a Google PhD fellowship.

\bibliographystyle{acl_natbib}
\bibliography{references}

\clearpage
\appendix\section*{Appendix}

\paragraph{Datasets Size}

\begin{table}[b]
\centering
\resizebox{0.8\columnwidth}{!}{
\begin{tabular}{@{}llccc@{}}
\toprule
Dataset                   & Split     & train & dev & test \\
\midrule
\multirow{4}{*}{CFQ}      & iid       & 95K   & 12K & 12K  \\
                          & MCD1      & 95K   & 12K & 12K  \\
                          & MCD2      & 95K   & 12K & 12K  \\
                          & MCD3      & 95K   & 12K & 12K  \\
\midrule
\multirow{2}{*}{ATIS}     & iid       & 4347  & 447 & 486  \\
                          & Template  & 4812  & 121 & 347  \\
                          \midrule
\multirow{2}{*}{GeoQuery} & iid       & 549   & 49  & 279  \\
                          & Template  & 539   & 159 & 182  \\
                          \midrule
\multirow{2}{*}{Scholar}  & iid       & 499   & 100 & 218  \\
                          & Template  & 408   & 94  & 315  \\
                          \midrule
\multirow{6}{*}{SCAN}     & iid       & 16782 & -   & 4182 \\
                          & Length    & 11990 & -   & 3920 \\
                          & Turn Left & 21890 & -   & 1208 \\
                          & MCD1      & 8365  & -   & 1045 \\
                          & MCD2      & 8365  & -   & 1045 \\
                          & MCD3      & 8365  & -   & 1045 \\
\bottomrule
\end{tabular}}
\caption{Sizes of all datasets and splits.}
\label{tab:dataset_size}
\end{table}

We include the sizes of all datasets and splits we experimented with in Table \ref{tab:dataset_size}.

\paragraph{Performance om i.i.d. Splits}

Full results are in Table~\ref{tab:iid_results}.

\paragraph{Training Details}
We fine-tune T5-Large and smaller on 32 Cloud TPU v3 cores, and use 128 cores for T5-3B.\footnote{https://cloud.google.com/tpu/}
Fine-tuning takes approximately 48 hours for T5-3B, and less than 16 hours for T5-Large and smaller.

\paragraph{Model Sizes}
We experiment with T5-small (60 million parameters), T5-base (220 million parameters), T5-large (770 million parameters) and T5-3B (3 billion parameters).

\begin{table*}[t]
\centering
\resizebox{0.6\textwidth}{!}{
\begin{tabular}{@{}lccccc@{}}
\toprule
           & \cfq{}  & \atis{} & \geo{}  & \scholar{} & \scan{} \\ \midrule
Baseline   & 99.5 & 58.6 & 80.6 & 72.9    & 100.0  \\
RIR        & 99.4 & 58.6 & 80.6 & 72.9    & 100.0  \\
\LIRdir     & 99.3 & 58.4 & 79.6 & 73.9    & 100.0  \\
\LIRdir+RIR & 99.3 & 59.3 & 79.2 & 74.3    & 100.0  \\
\LIRind     & 99.4 & 58.8 & 81.0 & 75.2    & 100.0  \\
\LIRind+RIR & 99.4 & 59.5 & 80.6 & 72.9    & 100.0  \\ \bottomrule
\end{tabular}}
\caption{Results on the test set for the for all approaches and all i.i.d. splits with T5-base.}
\label{tab:iid_results}
\end{table*}

\end{document}